\documentclass[letterpaper]{article} 
\usepackage{aaai2026}  
\usepackage{times}  
\usepackage{helvet}  
\usepackage{courier}  
\usepackage[hyphens]{url}  
\usepackage{graphicx} 
\usepackage{natbib}  
\usepackage{caption} 
\usepackage{algorithm}
\usepackage{algorithmic}

\usepackage{amsmath}
\usepackage{booktabs}



\usepackage{newfloat}
\usepackage{listings}
\DeclareCaptionStyle{ruled}{labelfont=normalfont,labelsep=colon,strut=off} 
\floatstyle{ruled}
\newfloat{listing}{tb}{lst}{}
\floatname{listing}{Listing}
\title{HLG: Comprehensive 3D Room Construction via Hierarchical Layout Generation}

\author{
    Xiping Wang\textsuperscript{\rm 1,\rm 4}\thanks{Equal contribution.}\,,\;
    Yuxi Wang\textsuperscript{\rm 2}\footnotemark[1]\,,\;
    Mengqi Zhou\textsuperscript{\rm 1,\rm 4},\;
    Junsong Fan\textsuperscript{\rm 3},\;
    Zhaoxiang Zhang\textsuperscript{\rm 1,\rm 3,\rm 4}\thanks{Corresponding author.}
}
\affiliations {
    \textsuperscript{\rm 1}University of Chinese Academy of Sciences (UCAS)\\
    \textsuperscript{\rm 2}Ocean University of China (OUC)\\
    \textsuperscript{\rm 3}Centre for Artificial Intelligence and Robotics (CAIR)\\
    \textsuperscript{\rm 4}Institute of Automation, Chinese Academy of Sciences (CASIA)\\
    \{wangxiping2024, zhoumengqi2022, junsong.fan, zhaoxiang.zhang\}@ia.ac.cn, yuxiwang93@gmail.com
}

\usepackage{bibentry}

\begin{document}
\twocolumn[{
\renewcommand\twocolumn[1][]{#1}
\maketitle

\begin{center}
    \vspace{-0.6cm}
    \centering
    \captionsetup{type=figure}
    \includegraphics[width=0.96\textwidth]{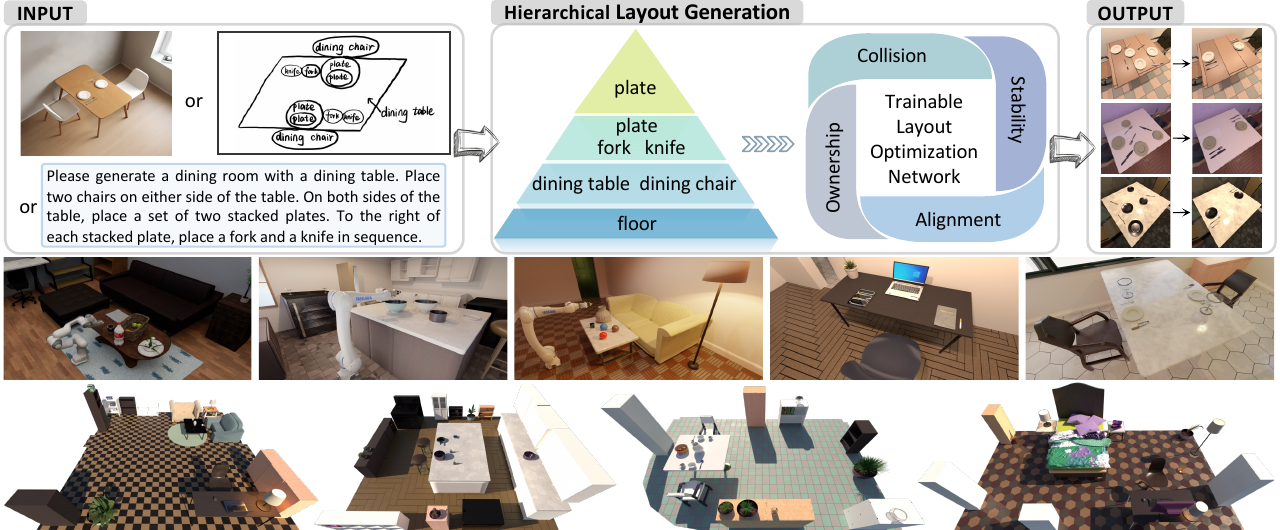}
    \caption{The proposed Hierarchical Layout Generation (HLG) can generate fine-grained 3D indoor scenes from multimodal inputs, including images, sketches, and test descriptions. Our method facilitates applications in virtual reality, interior design, and embodied intelligence, enabling the creation of highly detailed and realistic 3D environments.}
    \label{fig:Fig1}
\end{center}
}]

\footnotetext{\textsuperscript{*}Equal contribution}
\footnotetext{\textsuperscript{\dag}Corresponding author}

\begin{abstract}
Realistic 3D indoor scene generation is crucial for virtual reality, interior design, embodied intelligence, and scene understanding. 
While existing methods have made progress in coarse-scale furniture arrangement, they struggle to capture fine-grained object placements, limiting the realism and utility of generated environments. This gap hinders immersive virtual experiences and detailed scene comprehension for embodied AI applications.
%
To address these issues, we propose Hierarchical Layout Generation (\textbf{\textit{HLG}}), a novel method for fine-grained 3D scene generation. HLG adopts a coarse-to-fine hierarchical approach, refining scene layouts from large-scale furniture placement to intricate object arrangements. Specifically, our fine-grained layout alignment module constructs a hierarchical layout through vertical and horizontal decoupling, effectively decomposing complex 3D indoor scenes into multiple levels of granularity. Additionally, our trainable layout optimization network addresses placement issues, such as incorrect positioning, orientation errors, and object intersections, ensuring structurally coherent and physically plausible scene generation.
We demonstrate the effectiveness of our approach through extensive experiments, showing superior performance in generating realistic indoor scenes compared to existing methods. This work advances the field of scene generation and opens new possibilities for applications requiring detailed 3D environments. We will release our code upon publication to encourage future research.

\end{abstract}

\section{Introduction}
\label{sec:intro}



3D indoor scene generation is a critical area of research, with applications spanning virtual reality~\cite{wohlgenannt2020virtual}, interior design~\cite{FurniScene}, embodied intelligence~\cite{ma2024survey, OpenVLA}, and scene understanding~\cite{GGSD}.
Recent datasets, such as FurniScene~\cite{FurniScene}, have emerged to provide a large-scale collection of rooms with intricate furnishings and rich details, which are crucial for advancing these applications.
These extensive applications require 3D layout generation to feature immersive environments, accurate spatial representations, and precise visualization. Consequently, advancing 3D indoor scene generation not only contributes to the development of these fields but also plays a crucial role in building intelligent systems capable of reasoning, planning, and interacting in complex environments. 


Recent advancements in 3D indoor scene generation have led to the development of two main categories of methods: generation-based methods~\cite{Text2Room, Text2Scene, ATISS, DiffuScene, InstructScene, GenRC} and Large Language Model (LLM)-based methods~\cite{Holodeck, LayoutGPT, Chat2Layout, LLplace}. The first category focuses on generative models that directly synthesize 3D layouts from various inputs. For example, Text2Room \cite{Text2Room} and Text2Scene \cite{Text2Scene} generate 3D scene meshes through textual input. GenRC \cite{GenRC} produces a 3D room completion from spare images. Moreover, ATISS \cite{ATISS}, DiffuScene \cite{DiffuScene}, and InstructScene \cite{InstructScene} employ VAE structures and diffusion models to obtain 3D room layouts. 
The second category leverages large language models (LLMs, \textit{e.g.} GPT-4 \cite{GPT-4}) to generate 3D environments according to corresponding text prompts, such as LayoutGPT \cite{LayoutGPT}, Holodeck \cite{Holodeck}, LLplace \cite{LLplace}, and Chat2Layout \cite{Chat2Layout}. Different from generative methods that use various spatial feature priors to generate 3D layouts, LLM-based methods depend on language understanding to process textual prompts and convert them into spatial structures.
In addition, LLM-driven procedural generated content (PCG) methods have attracted much attention recently. For example, SceneX \cite{SceneX}, SceneCraft \cite{SceneCraft}, CityX \cite{CityX}, and RoomCraft \cite{RoomCraft} propose an automatic framework to generate large-scale environments using LLM agents. 
While these approaches show promising capability in creating high-level spatial relations, they still face significant challenges, particularly in generating fine-grained layouts. For instance, generating precise and meaningful arrangements of smaller objects, such as books, vases, or decorative items, remains a complex problem. This challenge is not just about filling space but also about creating natural, human-like placements that align with real-world usage and functionality.

To address these issues, our approach aims to generate fine-grained decoration placement, as shown in Fig.~\ref{fig:Fig1}. For example, at the table level, we focus on the precise placement of smaller objects like books, decor, and kitchenware. It is essential for interior design and embodied intelligence, as the realistic indoor environment ensures cohesive space visualization and provides diverse, high-quality datasets for tasks like object manipulation and navigation.


In this paper, we propose Hierarchical Layout Generation (HLG), a method for fine-grained 3D layout synthesis that introduces a hierarchical approach to refine indoor scene generation from coarse furniture arrangements to intricate object placements. By structuring the scene generation process hierarchically, HLG ensures that both large-scale spatial organization and fine-grained object positioning are handled in a coherent and controlled manner, leading to highly detailed and functionally realistic indoor environments.
%
Specifically, our fine-grained layout alignment module employs vertical and horizontal decoupling to systematically decompose complex 3D indoor scenes into multiple levels of granularity, ensuring precise object arrangement. Furthermore, our trainable layout optimization network actively corrects placement inconsistencies, such as misalignment, orientation errors, and object intersections, enhancing the structural coherence and physical plausibility of generated scenes.
By integrating these components, HLG bridges the gap between coarse layout generation and precise object placement, addressing key limitations of prior work and significantly advancing the realism and usability of 3D scene synthesis.


    
    
    

\section{Related Work}
\label{sec:related}

\subsection{3D Layout Generation}
Recent 3D layout generation methods can be categorized into two main paradigms: Generative Model-based Methods and LLM-based Methods.

\noindent\textbf{Generative Model-based Methods.} Early research~\cite{Total3DUnderstanding, ATISS, 3D-FRONT} models spatial distributions to synthesize realistic layouts while maintaining object relationships, but they lack explicit spatial reasoning. To improve physical plausibility, diffusion-based methods~\cite{DiffuScene, diffindscene, midi, gala3d, MVDiffusion} iteratively refine layouts, with cascaded sparse diffusion improving room-scale synthesis and enhancing 3D consistency. Building on advancements in diffusion models, HiScene~\cite{HiScene} generates hierarchical 3D scenes from isometric views in a top-down manner, focusing on amodal completion and spatial alignment for individual objects. Alternative approaches~\cite{PSDR-Room, LucidDreamer, roomdesigner, dreamscene, controlroom3d, huang2025unposed, debara, EchoScene} explore different representations to enhance scene coherence, integrating latent-based structure encoding, scene graph constraints, and score-based denoising for structured scene synthesis. 

\noindent\textbf{LLM-based Methods.} LLM-based methods~\cite{SceneCraft, SceneTeller, Chat2Layout, I-Design, LLplace, editroom, RoomCraft} prioritize natural language interaction for layout generation. LayoutGPT~\cite{LayoutGPT} employs in-context learning with CSS-style prompts to generate numerical layouts from text, while Holodeck~\cite{Holodeck} generates layouts by using search to solve constraints formed by relative relationships. LayoutVLM~\cite{LayoutVLM} combines the advantages of these two methods, utilizing a differentiable optimization framework to produce physically plausible and semantically coherent scenes. Recent hierarchical methods~\cite{AnyHome, 2025aaaihi} further organize scene generation into multiple levels. AnyHome\cite{AnyHome} introduces a two-level hierarchy (house-level and room-level), whereas the latter\cite{2025aaaihi} adopts a three-level structure with root, internal, and leaf nodes. However, these approaches lack fine-grained control over small object placement and struggle with handling non-rectangular rooms.

\begin{figure*}[!tbp]
  \centering
  \includegraphics[width=\textwidth]{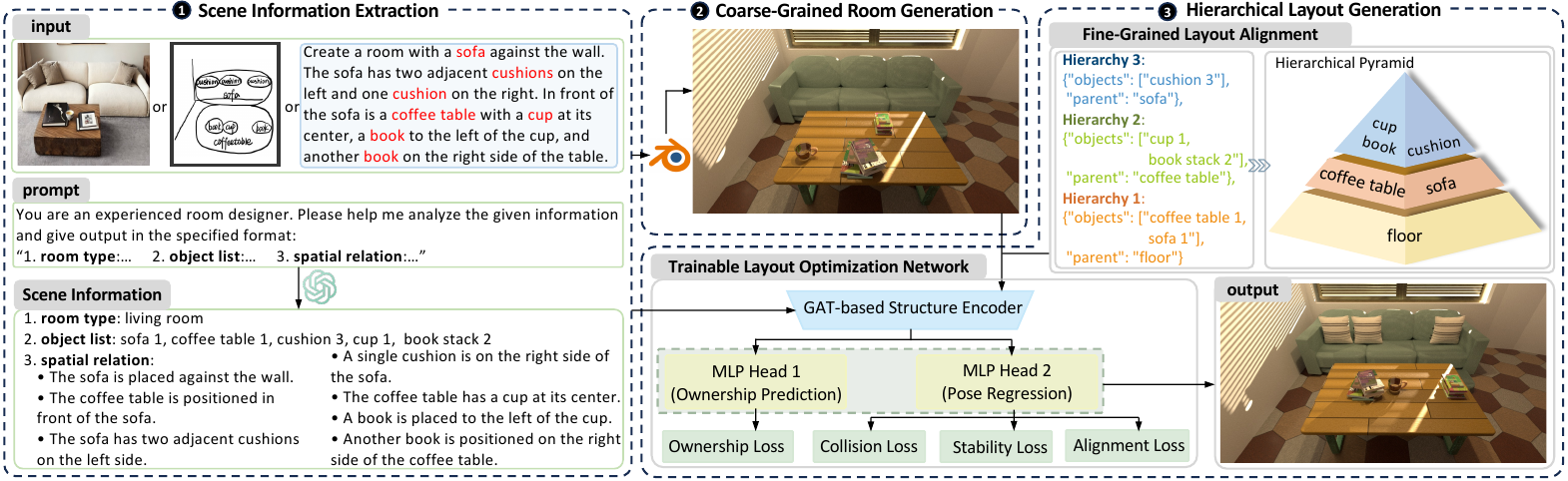}
  \caption{The proposed Hierarchical Layout Generation (HLG) framework generates a fine-grained 3D room layout based on multimodal instructions, including images, text descriptions, and sketches. It comprises three key stages: (1) scene information extraction, (2) coarse-grained room generation, and (3) hierarchical layout generation.}
 \label{fig:framework}
\end{figure*}

\subsection{Procedural Based Scene Generation}
Traditional Procedural Content Generation (PCG) methods~\cite{pmc, pcg_city, pcg_mpm, hendrikx2013procedural, ProcTHOR, pcg_plant} rely on rule-based templates to ensure deterministic outputs, while Infinigen~\cite{Infinigen} and its indoor variant~\cite{InfinigenIndoor} automate scene creation through predefined objects and architectural rules. These methods demonstrate strong stability in generation, producing collision-free and structurally predictable scenes. However, designing and modifying complex procedural rules requires substantial domain expertise. Recent PCG approaches aim to democratize controllability. 3DGPT~\cite{3D-GPT} translates textual prompts into procedural code, allowing users to specify styles. SceneX~\cite{SceneX} introduces a PCGHub where LLMs orchestrate asset generation and spatial relationships, enabling high-level directives. However, constrained by coarse-grained rule adaptation, these procedural methods also lack fine-grained spatial reasoning and adaptability, making it challenging to control the precise placement of small objects. To address this limitation, our HLG method introduces a hierarchical layout generation framework that ensures global structural consistency while enhancing the fine-grained arrangement of small objects.

\section{Method}
\label{sec:method}

\subsection{Pipeline Overview}
\label{subsec:pipeline}
The proposed Hierarchical Layout Generation (HLG) framework generates fine-grained 3D room layouts from multimodal inputs, including images, text descriptions, and sketches. It consists of three key stages: the \textit{scene information extraction} identifying key objects and spatial relationships, the \textit{coarse-grained room generation} constructing an initial 3D layout, and the \textit{hierarchical layout generation} refining object placements through proposed fine-grained layout alignment and optimization, shown in Fig.\ref{fig:framework}.

\noindent\textbf{Scene Information Extraction.} Inspired by the previous works~\cite{Holodeck, VisualHarmony, SceneX}, we first utilize GPT-4o~\cite{GPT-4o} acting as an agent to extract key information from input instructions (\textit{text, images, or hand-drawn sketches}). 
Specifically, it produces the essential information for layout generation, including the room type, furniture types, and approximate locations derived from the given prompts. 
Subsequently, it extracts more precise details required for fine-grained alignment, refining the relative positioning of furniture and accurately determining object placement within the room. 

\noindent\textbf{Coarse-Grained Room Generation.}
Existing works (\textit{e.g., SceneCraft~\cite{SceneCraft}, Chat2Layout~\cite{Chat2Layout}, LLplace~\cite{LLplace}, and VisualHarmony~\cite{VisualHarmony}}) have shown the potential of using procedural content generation software to create realistic 3D rooms in a retrieval and placement manner. Following these approaches, we propose the coarse-grained room generation module, integrating the basic room structure, such as walls, doors, and windows, and the approximate placement of furniture in a 3D space. At this stage, the overall room layout is initially completed, but the precise placement of furniture and detailed refinements will be further optimized in the subsequent steps. 

\noindent\textbf{Hierarchical Layout Generation.} 
In this stage, we refine object placement by incorporating hierarchical relationships, spatial constraints, and object interactions, ensuring spatial rationality and consistency beyond the coarse-grained room structure. This process organizes objects into a hierarchical structure with ownership relationships and employs a trainable layout optimization network to resolve collisions, enforce physical stability, and align objects with input specifications. By iteratively adjusting positions based on spatial constraints and a pretrained network, this stage ensures that objects maintain logical spatial relationships, avoid conflicts, and adhere to design constraints, resulting in a structured, functionally feasible, and aesthetically consistent room layout.

\subsection{Hierarchical Layout Generation}
\begin{figure}[!tbp]
    \centering
    \includegraphics[width=1\linewidth]{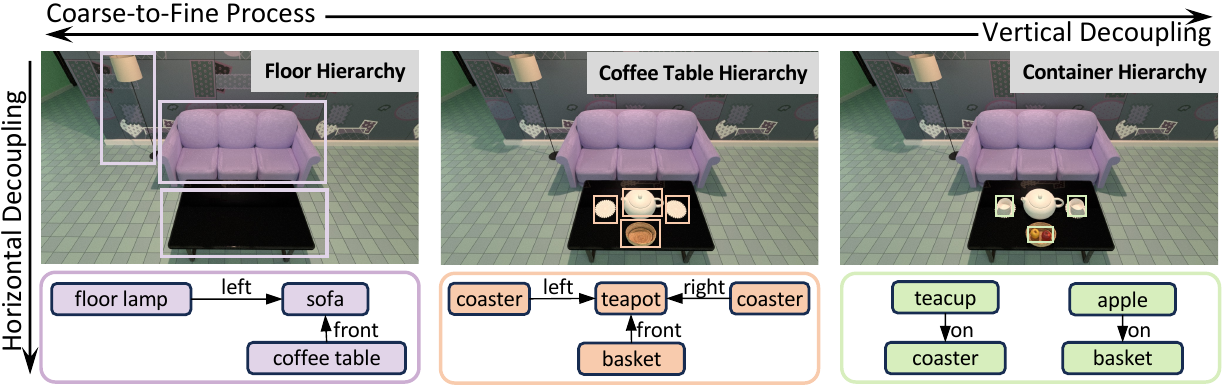}
    \caption{Illustration of Hierarchical Layout Generation. Floor Hierarchy for large furniture (\textit{e.g.}, sofa, coffee table and floor lamp), Coffee Table Hierarchy for tabletop objects (\textit{e.g.}, teapot, coaster and basket), and Container Hierarchy for fine-level placements (\textit{e.g.}, teacup and apple).}
    \label{fig:HLG}
\end{figure}

To represent a comprehensive 3D room scene $\mathcal{S}$, we assume it should compose the domain of room type, furniture in the room, and the relationship or arrangement for the object. Therefore, a room $\mathcal{S}$ can be represented by:
\begin{equation}
\mathcal{S}=(\mathcal{O},\mathcal{D},\mathcal{C}),
\end{equation}
where $\mathcal{O}$ denotes the objects in the scene (\textit{e.g.}, furniture or decorations), $\mathcal{D}$ represents the room types, such as living room, bedroom, \textit{etc.}, and $\mathcal{C}$ is the constraints for the objects, exampling as \{``\textit{face to}'', ``\textit{left to}'', ``\textit{right to}'', ``\textit{front}'' \textit{etc.}\}. To reduce the complexity of generating room $\mathcal{S}$, we propose a hierarchical layout generation method by dividing the layout into different hierarchies in a coarse-to-fine manner. For example, as Fig.~\ref{fig:HLG} shows, a living room can be divided into three hierarchies, including the floor hierarchy, the coffee table hierarchy, and the container hierarchy. To achieve the fine-grained object placement, the proposed HLG includes \textit{fine-grained layout alignment} and \textit{trainable layout optimization network}.

\subsubsection{Fine-Grained Layout Alignment.}
This module aims to achieve fine-grained alignment by establishing a clear and accurate hierarchy through \textit{vertical decoupling} and \textit{horizontal decoupling}.

\textit{Vertical Decoupling.} The primary challenge in spatial layout generation lies in handling the complexity of hierarchical relationships between objects. Traditional methods often struggle with dependencies across different objects, leading to optimization bottlenecks and suboptimal solutions. To address this issue, we propose a vertical decoupling method, which decomposes a 3D scene into independent vertical layers along the Z-axis.
    
Specifically, each object is assigned to a single parent level based on its spatial relationship (\textit{e.g.}, ``teacup on the coffee table''). This creates a chain-like structure where each layer represents a fine-grained hierarchy and can be subsequently optimized. The decomposition adheres to the following constraint:
\begin{equation}
\mathcal{C}_i \cap \mathcal{C}_j = \emptyset, \quad \forall i \neq j ,
\end{equation}
where $\mathcal{C}_i$ and $\mathcal{C}_j$ represent constraints within different layers. By isolating these constraints, we avoid conflicts that arise from multiple parent-level associations, such as floating or clipped objects.

The resulting hierarchical structure enables modular optimization. For example, the layout of a table can be optimized independently while ensuring alignment with its parent (\textit{e.g.}, the floor). This approach significantly reduces the problem's complexity and improves the quality of generated layouts.

\textit{Horizontal Decoupling.}
While vertical decoupling simplifies optimization by isolating layers, handling spatial relationships within each layer remains non-trivial. To tackle this, we introduce horizontal decoupling, which leverages large language models (LLMs) to transform complex spatial relationships into structured graph data. Inspired by prior work~\cite{SceneCraft, Holodeck}, we utilize an LLM agent to construct a scene graph for each fine-grained hierarchy.
The scene graph is represented as nodes and edges, where nodes correspond to objects and edges encode their spatial relationships or constraints. 
Each hierarchy specifies positional constraints (\textit{e.g.}, ``teapot at the center of the table'', or ``coaster left to teapot''). This step ensures the LLM generates consistent and interpretable relationships, providing a solid foundation for subsequent layout generation.

By decomposing spatial relationships into structured graph data, horizontal decoupling enables efficient problem-solving using greedy algorithms. For instance, the scene graph can be processed level by level, starting from parent objects (\textit{e.g.}, furniture) and recursively placing child objects (\textit{e.g.}, items on the desk). This approach ensures alignment between parent and child objects while gradually refining the layout from coarse-grained to fine-grained structures.

\subsubsection{Trainable Layout Optimization Network.}
While prior works rely on rule-based constraints to determine object placements, we introduce a Trainable Layout Optimization Network (TLO-Net), which refines object positions and orientations in a differentiable manner. TLO-Net consists of a structure encoder based on Graph Attention Network (GAT)~\cite{gat} and two task-specific multilayer perceptron (MLP)~\cite{mlp} heads responsible for hierarchical ownership prediction and fine-grained pose regression, respectively.

TLO-Net takes three inputs: (i) the input instructions $\mathcal{I}$, (ii) the initial scene layout \( \mathcal{S}_{coarse} \) generated by the coarse-grained room generation module, and (iii) the hierarchical structure graph $\mathcal{G}$ generated by the fine-grained layout alignment module. To learn to generate physically plausible and instruction-aligned layouts, TLO-Net processes these inputs and optimizes its parameters by minimizing a composite loss function, defined as:
\begin{align}
\mathcal{L}_{\mathrm{total}} ={}& \mathcal{L}_{\mathrm{owner}} + \mathcal{L}_{\mathrm{collision}} + \mathcal{L}_{\mathrm{stability}} + \mathcal{L}_{\mathrm{align}}
\end{align}
Each component is detailed as follows:

\begin{figure*}[!tbp]
    \centering
    \includegraphics[width=0.98\linewidth]{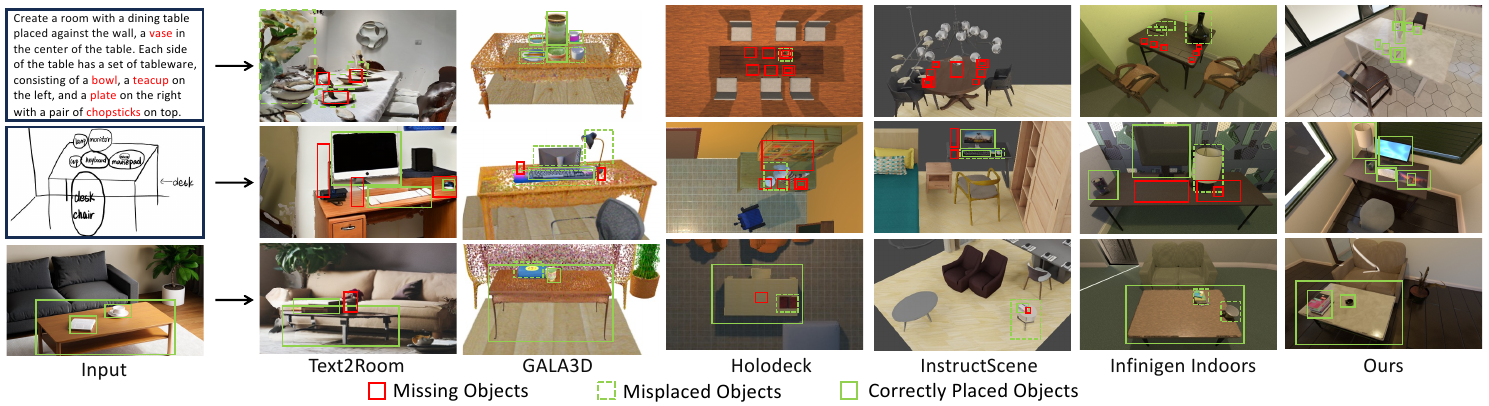}
    \caption{Comparison of fine-grained desktop layout generation. While existing methods struggle with object completeness, spatial accuracy, and adherence to input constraints, HLG effectively preserves input details and ensures structured, fine-grained object placement with high controllability.}
    \label{fig:exp1}
\end{figure*}

\textit{Ownership Loss~($\mathcal{L}_{\mathrm{owner}}$)}: To ensure each object is correctly assigned to exactly one parent within the hierarchical structure, we define the ownership loss as a supervised classification objective:
\begin{equation}
\mathcal{L}_{\mathrm{owner}}=-\sum_{o_p\in\mathcal{O}_i, o_q \in \mathcal{O}_j} g_{pq}\text{log}(p_{pq}),
\end{equation}
where $o_p$ denotes a child object and $o_q$ denotes a potential parent object for $o_p$. The summation iterates over all child objects and their respective sets of potential parents. $g_{pq}$ is 1 if $o_q$ is the true parent of $o_p$ and 0 otherwise, and $p_{pq}$ is TLO-Net's predicted probability that $o_q$ is the parent of object $o_p$.
    
\textit{Collision Loss~($\mathcal{L}_{\mathrm{collision}}$)}: To prevent unwanted physical overlaps between objects in the generated 3D scene, we define a collision loss that penalizes 3D bounding box intersections as follows:
\begin{equation}
\mathcal{L}_{\mathrm{collision}}=\sum_{o_p,o_q\in \mathcal{O}, o_p \neq o_q} \text{IoU}(\ \mathrm{BBOX}(o_p), \mathrm{BBOX}(o_q)),
\end{equation}
where $\mathrm{BBOX}(o)$ denotes the 3D bounding box of object $o$, and intersection-over-union (IoU) measures the volumetric overlap between two bounding boxes, with higher IoU values indicating greater spatial conflict.
    
\textit{Stability Loss~($\mathcal{L}_{\mathrm{stability}}$)}: To prevent physically implausible placements such as tipping or floating, we define a stability loss that penalizes objects whose centers of mass (COM) fall outside their support polygon. Specifically:
\begin{equation}
\mathcal{L}_{\mathrm{stability}}=\sum_{o\in\mathcal{O}} \text{dist}(\text{Proj}_z(\text{COM}(o)), \text{Support}(o)), 
\end{equation}
where $\text{Proj}_z(\text{COM}(o))$ is the projection of object $o$'s COM onto the horizontal plane, and $\text{Support}(o)$ denotes its corresponding support polygon. The function $\text{dist}(\cdot,\cdot)$ calculates the minimum Euclidean distance from the projected COM to the boundary of the support polygon, and is defined as zero if the point lies within the boundary.
    
\textit{Alignment Loss~($\mathcal{L}_{\mathrm{align}}$)}: To minimizes the difference between the generated 3D scene \( \mathcal{S} \) and user-defined instructions $\mathcal{I}$, we define the alignment loss as follows:
\begin{equation}
\mathcal{L}_{\mathrm{align}}=\sum_{o\in\mathcal{O}}\|\mathcal{P}(o)-\mathcal{P}_i(o)\|^2, 
\end{equation}
where $\mathcal{P}(o)$ indicates the position and orientation of object $o$ in the generated scene, and the $\mathcal{P}_i(o)$ is the corresponding position and orientation from the input instruction. It should notice that the $\mathcal{P}(o)$ can be obtained from the software (\textit{e.g., Blender}), and the $\mathcal{P}_i(o)$ is initialized by the LLM. 

We train the TLO-Net using synthetic layouts with annotated positions and relationships. The model iteratively updates the layout in a differentiable manner, allowing gradients to backpropagate from downstream losses. At inference time, the same model can be applied iteratively or in a single pass to refine coarse layouts into fine-grained results.

\section{Experiments}
\label{sec:exp}

\subsection{Benchmark Protocol}
\label{subsec:Benchmark}

\begin{figure*}[!tbp]
    \centering
    \includegraphics[width=0.96\linewidth]{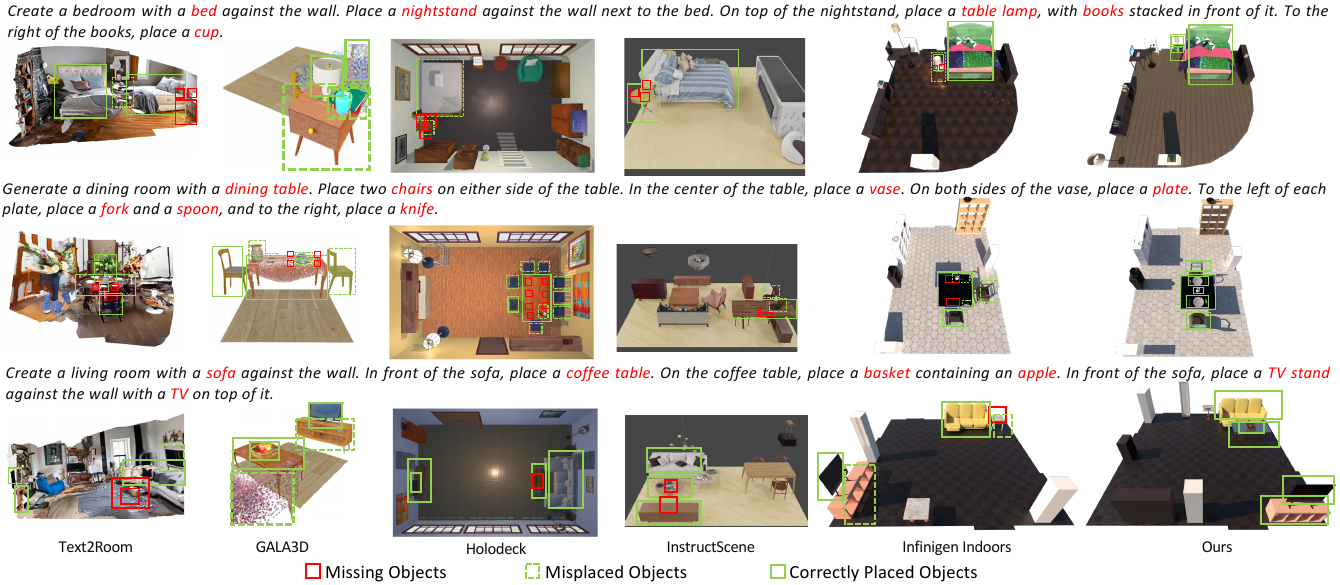}
    \caption{Comparison of complete room layout generation. HLG is evaluated against five baseline methods. While existing methods struggle with missing furniture, geometric inconsistencies, or limited scene details, HLG effectively balances global layout structure with fine-grained object placement, ensuring spatial coherence and input alignment.}
    \label{fig:exp2}
\end{figure*}

\noindent\textbf{Dataset.} We construct a dataset consisting of 150 room scenes, encompassing three input modalities: 50 instructions generated by GPT-4o~\cite{GPT-4o}, providing detailed descriptions of object placements; 50 user-drawn 2D sketches, annotating key object positions and their spatial relationships; and 50 RGB images collected from interior design websites. Each scene contains at least five objects, with fine-grained objects (\textit{e.g.}, decorations and household items) comprising over 30\%. For fair comparison, we adapt each scene to baseline input formats.

\noindent\textbf{Baselines.} We evaluate five representative 3D room generation methods, covering key technological paradigms such as image-based fusion, generative models, LLM-driven approaches, graph-conditioned diffusion models, and procedural content generation(PCG). 
\begin{itemize}
    \item \textbf{Text2Room}~\cite{Text2Room} iteratively integrates 2D text-to-image outputs with monocular depth estimation for 3D scene meshes;
    \item \textbf{GALA3D}~\cite{gala3d}, which generates an initial layout using an LLM and refines a 3D Gaussian representation via a diffusion model; 
    \item \textbf{Holodeck}~\cite{Holodeck}, an LLM-driven procedural generation approach that leverages LLMs to generate 3D indoor scenes from user instruction automatically;
    \item \textbf{InstructScene}~\cite{InstructScene}, which models object layouts and generates indoor scenes through semantic graph prior and diffusion model; 
    \item \textbf{Infinigen Indoors}~\cite{InfinigenIndoor}, a fully procedural indoor scene generation framework that defines layout rules through a constraint language. 
\end{itemize}
As GALA3D~\cite{gala3d} is not publicly available, we reimplement it based on the details provided in the paper. Due to the lack of details, the results may not fully match the original performance. In addition, several recent methods are either not open-sourced (\textit{e.g.}, HiScene~\cite{HiScene}, CASAGPT~\cite{CASAGPT}, and other LLM-based methods~\cite{2025aaaihi, 20253dagent}), partially released (\textit{e.g.}, LayoutVLM~\cite{LayoutVLM}, FlairGPT~\cite{FlairGPT}), or have incorrect code (\textit{e.g.}, MMGDreamer~\cite{MMGDreamer}). As a result, we are unable to include them in our comparison.

\noindent\textbf{Metrics.} Inspired by previous work~\cite{Chat2Layout}, we adopt four complementary metrics to evaluate layout quality and input consistency. \textit{Out-of-Bound Rate (OOB)} quantifies physical plausibility by measuring the percentage of objects exceeding room boundaries or parent support surfaces. \textit{Orientation Correctness (ORI)} assesses the accuracy of angular alignment. \textit{CLIP-Similarity (CLIP-Sim)} evaluates the alignment between user instructions and scene content. Finally, \textit{Fréchet Inception Distance (FID)~\cite{fid}} measures the quality of images created by generative models. 

\subsection{Qualitative Experimental Results}
\label{subsec:Experimental Results}
\noindent\textbf{Fine-Grained Desktop Layout.} To validate the ability of HLG to achieve fine-grained small object placement, we conduct an experimental evaluation on the desktop hierarchy. As shown in Fig.~\ref{fig:exp1}, compared to baseline methods, our method demonstrates significant advantages in object completeness and positional accuracy.

Specifically, Text2Room~\cite{Text2Room} struggles to meet input constraints, as the generated desktop not only fails to include key objects such as teacups and desk lamps, but also exhibits severe spatial misalignment of existing objects (\textit{e.g.}, all objects on the dining table deviate significantly from the input description). 
GALA3D~\cite{gala3d} heavily relies on LLM tuning and parameter configurations, leading to unreasonable spatial arrangements of small objects. 
Holodeck~\cite{Holodeck} fails to generate fine-grained desktop layouts, with all expected items either misplaced or absent, indicating its limited capability in handling detailed constraints. 
InstructScene~\cite{InstructScene} suffers from massive small object omissions, reflecting the instability of diffusion models in generating fine-grained object placements. 
Infinigen Indoors~\cite{InfinigenIndoor}, despite leveraging procedural generation, lacks systematic planning, resulting in randomly placed small objects that ignore input constraints.
In contrast, our HLG framework maintains input fidelity and accurately generates fine-grained layouts via a hierarchical strategy, significantly enhancing controllability and spatial precision.

\noindent\textbf{Complete Room Layout.} To validate HLG's ability to generate complete room layouts that align with input constraints, we conducted experiments on full-room scene generation and compared our approach with baseline methods. 

As shown in Fig.~\ref{fig:exp2}, Text2Room~\cite{Text2Room} produces visually realistic scenes but exhibits weak geometric constraint modeling and often omits key furniture such as nightstand and coffee table. 
GALA3D~\cite{gala3d}, relying on coordinate-based layout generation, is highly dependent on LLM guidance and parameter tuning. It requires well-optimized coordinates, angles, and object parameters from the LLM to achieve reasonable layouts. 
Holodeck~\cite{Holodeck} generates scenes based on common furniture configuration patterns, but consequently struggles to fully align with specific user instructions. 
InstructScene~\cite{InstructScene} performs well in large-scale furniture arrangement (\textit{e.g.}, sofas and beds). However, severe geometry penetration issues are observed (\textit{e.g.}, dining chairs intersecting the dining table), and the generated scenes are highly simplified, lacking nearly all fine-grained objects, thereby limiting scene richness. 
Infinigen Indoors~\cite{InfinigenIndoor}, despite ensuring basic furniture presence via procedural rules, fails to fully align with input constraints, leading to suboptimal layout rationality (\textit{e.g.}, the TV stand in front of the sofa is missing despite being specified in the input).
In contrast, HLG explicitly models spatial hierarchies, balancing accurate large furniture placement with fine-grained detail, demonstrating superior global structure and input alignment.  

\subsection{Quantitative Experimental Results}
\begin{table}[!tbp]
\centering
\tabcolsep=4pt 
\small 
\begin{tabular}{lcccc}
\toprule
\textbf{Method} & \textbf{OOB}$\downarrow$ & \textbf{ORI}$\uparrow$ & \textbf{CLIP-Sim}$\uparrow$ & \textbf{FID}$\downarrow$ \\
\midrule
Text2Room & 35.5 & 57.3 & 18.9 & 50.2 \\
GALA3D & 30.1 & 66.5 & 27.2 & 45.9 \\
Holodeck & 63.3 & 44.8 & 13.5 & 40.2 \\
InstructScene & 36.7 & 78.2 & 13.6 & 42.3 \\
Infinigen Indoors & 21.6 & 75.3 & 27.1 & 40.6 \\
\midrule
Ours (w/o FGLA) & 18.1 & 87.2 & 28.1 & 39.7 \\
Ours (w/o $\mathcal{L}_{\mathrm{owner}}$) & 16.3 & 88.9 & 28.8 & 38.9 \\
Ours (w/o $\mathcal{L}_{\mathrm{collision}}$) & 17.1 & 89.0 & 28.2 & 40.3 \\
Ours (w/o $\mathcal{L}_{\mathrm{stability}}$) & 20.8 & 86.7 & 29.1 & 40.0 \\
Ours (w/o $\mathcal{L}_{\mathrm{align}}$) & 16.8 & 84.2 & 27.7 & 39.8 \\
Ours (w/o TLO-Net) & 20.9 & 84.3 & 27.5 & 40.1 \\
\midrule
\textbf{Ours} & \textbf{16.2} & \textbf{89.2} & \textbf{29.3} & \textbf{38.2} \\
\bottomrule
\end{tabular}
\caption{Layout quality comparison. Our method achieves superior performance with a lower OOB, better ORI, higher CLIP-Sim, and lower FID.}
\label{tab:OOB}
\end{table}
\noindent\textbf{Layout Quality Results.} 
As shown in Table~\ref{tab:OOB}, HLG consistently outperforms all baselines across four metrics. Specifically, HLG achieves the lowest OOB and highest ORI, indicating better spatial constraint adherence and object orientation accuracy. The highest CLIP-Sim score demonstrates stronger semantic alignment, and the lowest FID signifies improved realism and structural coherence. These results validate HLG's effectiveness in producing structured and semantically accurate room layouts.

\begin{figure}[!tbp]
    \centering
    \includegraphics[width=0.98\linewidth]{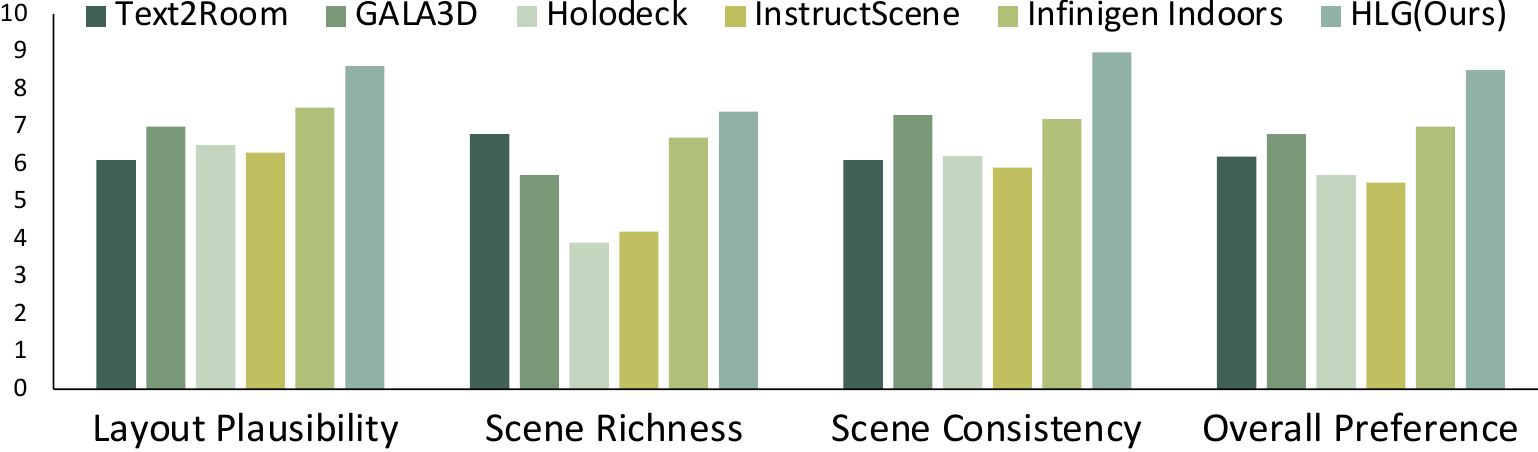}
    \caption{User Study Results on Layout Plausibility, Scene Richness, Scene Consistency, and Overall Preference.}
    \label{fig:userstudy}
\end{figure}

\noindent\textbf{User Study.} 
To further assess scene quality, we conducted a user study with 136 graduate students, who rated our method and five baselines on Layout Plausibility, Scene Richness, Scene Consistency, and Overall Preference. 
As shown in Fig.~\ref{fig:userstudy}, Text2Room~\cite{Text2Room} shows poor consistency due to misaligned layout. GALA3D~\cite{gala3d} exhibits moderate overall performance, as its dependence on LLMs leads to layouts that contain relevant objects but lack precise spatial organization. Holodeck's~\cite {Holodeck} low scene richness suggests that its LLM-driven object selection and placement might lead to less diverse scenes or difficulties in fine-grained object arrangements. InstructScene~\cite{InstructScene} lacks scene richness due to small object omissions. Infinigen Indoors~\cite{InfinigenIndoor} shows low consistency due to limited control over fine-grained layout and occasional misplacement of major furniture. In contrast, HLG surpasses all baselines across all metrics, affirming its ability to generate structured, rich, and spatially coherent scenes.

\subsection{Ablation Study}
\label{subsec:Ablation Study}
This section presents an ablation study to validate the critical roles of the Fine-Grained Layout Alignment (FGLA) module and the Trainable Layout Optimization Network (TLO-Net) in enhancing scene generation quality.

\begin{figure}[!tbp]
    \centering
    \includegraphics[width=0.98\linewidth]{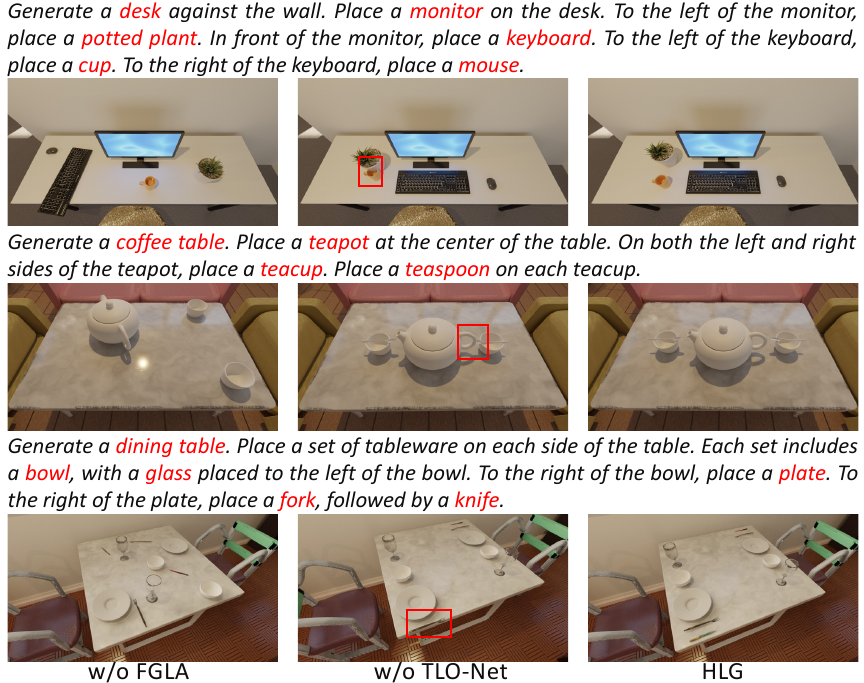}
    \caption{Visual results of ablation studies. Experiments evaluate the role of FGLA and TLO-Net, highlighting their necessity for scene consistency and physical rationality.}
    \label{fig:abl}
\end{figure}

\noindent\textbf{The influence of FGLA.} To evaluate the impact of FGLA, we remove it and conduct desktop scene generation experiments. The ablated model shows a notable degradation in spatial organization and semantic alignment capabilities. CLIP-Sim drops by 4.1\% (29.3 → 28.1), indicating reduced adherence to input constraints. Visual analysis (Fig.~\ref{fig:abl}) shows that objects lack structured placement, with missing hierarchies (\textit{e.g.}, absent teaspoon on teacup) and arbitrary positioning (\textit{e.g.}, misplaced keyboard and mouse). In contrast, HLG preserves vertical relationships, ensuring structured layouts and preserving fine-grained input details.

\noindent\textbf{The influence of TLO-Net.} As shown in Table~\ref{tab:OOB}, disabling TLO-Net increases OOB by 29.0\% (16.2\% → 20.9\%) and drops ORI by 5.5\% (89.2\%  → 84.3\% ), highlighting the degradation boundary handling and orientation correction. As shown in Fig.~\ref{fig:abl}, the ablated model fails to solve collision (\textit{e.g.}, a potted plant overlaps with a cup; a teaspoon penetrates a teapot), while also lacking physical stability enforcement, leading to objects appearing unbalanced or floating (\textit{e.g.}, floating tableware). 
In contrast, HLG ensures collision-free and physically consistent layouts, confirming its necessity for fine-grained scene optimization.

\section{Conclusion}
\label{sec:conclusion}
In this paper, we introduce Hierarchical Layout Generation (HLG), a novel framework for fine-grained 3D indoor scene synthesis that refines layouts from coarse furniture arrangements to precise object placement. HLG integrates a hierarchical generation strategy with a trainable layout optimization network, mitigating issues like misalignment, orientation errors, and object collisions, ensuring plausible scenes. Extensive experiments demonstrate HLG's superior spatial accuracy and realism, particularly in small object placement, advancing 3D scene generation for VR, interior design, and embodied AI applications.

\bibliography{aaai2026}
\end{document}